\documentclass[letterpaper]{article}
\usepackage{uai2019}
\usepackage[margin=1in]{geometry}

\usepackage{times}
\usepackage{microtype,marvosym}

\usepackage{subcaption}
\usepackage{amsmath,amsfonts,amssymb}
\usepackage[utf8]{inputenc}
\usepackage[T1]{fontenc}
\usepackage{lmodern}
\usepackage{nicefrac}
\usepackage{etoolbox}
\usepackage{enumitem}
\usepackage{mathtools}
\usepackage{extarrows}
\usepackage{float} 

\usepackage[super]{nth} 

\usepackage{xcolor}
\usepackage[hidelinks]{hyperref}

\definecolor{mred}{RGB}{200,0,25} 
\definecolor{forestgreen}{RGB}{34, 139, 34}

\newcommand{\rz}{Z} 
\newcommand{\gp}{\text{\textsc{gp}}}
\newcommand{\icm}{\text{\textsc{icm}}}
\newcommand{\bq}{{\sc bq}} 
\newcommand{\vbq}{\text{\sc vbq}} 
\newcommand{\amsbq}{\text{\sc ams-bq}} 
\newcommand{\mi}{\text{\sc mi}} 
\newcommand{\ivr}{\text{\sc ivr}} 
\newcommand{\ip}{\text{\sc ip}} 
\newcommand{\ldat}{\boldsymbol{\ell}}
\newcommand{\intangle}[1]{\langle #1 \rangle}
\newcommand{\iintangle}[1]{\langle\!\langle #1 \rangle\!\rangle}
\newcommand{\fvec}{\boldsymbol{f}}
\newcommand{\fint}{\langle f_1 \rangle}

\newcommand{\ode}{\textsc{ode}}
\newcommand{\suppmat}{supp.~mat.}

\newcommand{\ie}{i.e., } %
\newcommand{\eg}{e.g., } %
\newcommand{\g}{\,|\,}

\newcommand{\eps}{\epsilon}
\newcommand{\dd}{\:\mathrm{d}}

\newcommand{\Exp}{\mathbb{E}}

\newcommand{\cov}{\operatorname{cov}}

\renewcommand{\Re}{\mathbb{R}}

\newcommand{\one}{\mathbf{1}}

\newcommand{\diag}{\operatorname{diag}}
\newcommand{\Dcal}{\mathcal{D}} 
\newcommand{\N}{\mathcal{N}}

\renewcommand{\L}{\mathcal{L}}
\newcommand{\inv}{^{-1}} 
\newcommand{\Trans}{^{\intercal}} 

\newcommand{\argmax}{\operatorname*{arg\:max}}

\newcommand{\V}{\mathbb{V}}

\renewcommand{\vec}{\boldsymbol}

\newcommand{\GP}{\mathcal{GP}}
\newcommand{\Id}{\boldsymbol{{I}}}

\renewcommand{\L}{\mathcal{L}}

\newcommand{\f}{\boldsymbol{\mathsf{f}}}
\newcommand{\h}{\boldsymbol{\mathsf{h}}}

\newcommand{\kvec}{\boldsymbol{\mathsf{k}}}
\newcommand{\m}{\boldsymbol{\mathsf{m}}}
\renewcommand{\u}{\boldsymbol{\mathsf{u}}}

\newcommand{\x}{\boldsymbol{\mathsf{x}}}
\newcommand{\y}{\boldsymbol{\mathsf{y}}}

\newcommand{\sA}{\boldsymbol{\mathsf{A}}}
\newcommand{\sB}{\boldsymbol{\mathsf{B}}}

\newcommand{\sG}{\boldsymbol{\mathsf{G}}}
\newcommand{\sH}{\boldsymbol{\mathsf{H}}}

\newcommand{\sK}{\boldsymbol{\mathsf{K}}}

\newcommand{\sV}{\boldsymbol{\mathsf{V}}}
\newcommand{\sW}{\boldsymbol{\mathsf{W}}}
\newcommand{\sX}{\boldsymbol{\mathsf{X}}}
\newcommand{\sY}{\boldsymbol{\mathsf{Y}}}
\newcommand{\sSigma}{\boldsymbol{\mathsf{\Sigma}}}

\usepackage{colonequals}

\DeclareSymbolFont{stmry}{U}{stmry}{m}{n}
\DeclareMathSymbol\leftarrowtriangle\mathrel{stmry}{"5E}
\DeclareMathSymbol\rightarrowtriangle\mathrel{stmry}{"5F}
\DeclareMathSymbol\leftrightarrowtriangle\mathrel{stmry}{"5D}
\DeclareMathSymbol\obar\mathrel{stmry}{"3A}
\DeclareMathSymbol\otimes\mathrel{stmry}{"0F}
\DeclareMathSymbol\ominus\mathrel{stmry}{"17}
\DeclareMathSymbol\sslash\mathrel{stmry}{"0C}
\DeclareUnicodeCharacter{200E}{}

\setlist{itemsep=3pt}

\usepackage[style=authoryear,
            maxcitenames=2,
            maxbibnames=5,
            uniquelist=false,
            backend=biber,
            natbib=true, 
            url=false,
            doi=false,
            eprint=false,
            isbn=false,
            dashed=false,
            hyperref=true]{biblatex}
\DeclareCiteCommand{\cite}
  {\usebibmacro{prenote}}
  {\usebibmacro{citeindex}%
   \printtext[bibhyperref]{\usebibmacro{cite}}}
  {\multicitedelim}
  {\usebibmacro{postnote}}

\DeclareCiteCommand*{\cite}
  {\usebibmacro{prenote}}
  {\usebibmacro{citeindex}%
   \printtext[bibhyperref]{\usebibmacro{citeyear}}}
  {\multicitedelim}
  {\usebibmacro{postnote}}

\DeclareCiteCommand{\parencite}[\mkbibparens]
  {\usebibmacro{prenote}}
  {\usebibmacro{citeindex}%
    \printtext[bibhyperref]{\usebibmacro{cite}}}
  {\multicitedelim}
  {\usebibmacro{postnote}}

\DeclareCiteCommand*{\parencite}[\mkbibparens]
  {\usebibmacro{prenote}}
  {\usebibmacro{citeindex}%
    \printtext[bibhyperref]{\usebibmacro{citeyear}}}
  {\multicitedelim}
  {\usebibmacro{postnote}}

\DeclareCiteCommand{\footcite}[\mkbibfootnote]
  {\usebibmacro{prenote}}
  {\usebibmacro{citeindex}%
  \printtext[bibhyperref]{ \usebibmacro{cite}}}
  {\multicitedelim}
  {\usebibmacro{postnote}}

\DeclareCiteCommand{\footcitetext}[\mkbibfootnotetext]
  {\usebibmacro{prenote}}
  {\usebibmacro{citeindex}%
   \printtext[bibhyperref]{\usebibmacro{cite}}}
  {\multicitedelim}
  {\usebibmacro{postnote}}

\DeclareCiteCommand{\textcite}
  {\boolfalse{cbx:parens}}
  {\usebibmacro{citeindex}%
   \printtext[bibhyperref]{\usebibmacro{textcite}}}
  {\ifbool{cbx:parens}
     {\bibcloseparen\global\boolfalse{cbx:parens}}
     {}%
   \multicitedelim}
  {\usebibmacro{textcite:postnote}}

\DeclareFieldFormat{titlecase}{\MakeSentenceCase*{#1}}

\newcommand{\figpath}{.}

\newcommand{\lsim}{\raisebox{-0.13cm}{~\shortstack{$<$ \\[-0.07cm]
      $\sim$}}~}

\title{Active Multi-Information Source Bayesian Quadrature}

\author{ {\bf Alexandra~Gessner\thanks{\hspace{1ex}work primarily performed during an internship at Amazon Research, Cambridge, UK.}} \\
University of T\"ubingen \\
MPI for Intelligent Systems\\
T\"ubingen, Germany\\
\texttt{agessner@tue.mpg.de}\\
\And
{\bf Javier Gonzalez}  \\
Amazon Research          \\
Cambridge, UK \\
\texttt{gojav@amazon.com}\\
\And
{\bf Maren Mahsereci}   \\
Amazon Research  \\
Cambridge, UK   \\
\texttt{mahserec@amazon.com}\\
}

\begin{document}
\maketitle

\begin{abstract}
Bayesian quadrature (\bq) is a sample-efficient probabilistic numerical method to solve integrals of expensive-to-evaluate black-box functions, yet so far, \emph{active} \bq~learning schemes focus merely on the integrand itself as information source, and do not allow for information transfer from cheaper, related functions.
Here, we set the scene for active learning in \bq~when multiple related information sources of variable cost (in input and source) are accessible.
This setting arises for example when evaluating the integrand requires a complex simulation to be run that can be approximated by simulating at lower levels of sophistication and at lesser expense.
We construct meaningful cost-sensitive multi-source acquisition \emph{rates} as an extension to common utility functions from vanilla \bq~(\vbq), and discuss pitfalls that arise from blindly generalizing.
In proof-of-concept experiments we scrutinize the behavior of our generalized acquisition functions.
On an epidemiological model, we demonstrate that active multi-source \bq~(\amsbq) allocates budget more efficiently than \vbq~for learning the integral to a good accuracy.
\end{abstract}

\section{INTRODUCTION}
\label{sec:introduction}

Integrals of expensive-to-evaluate functions arise in many scientific and industrial applications, for example when expected values need to be computed and each evaluation of the integrand requires the run of a complex computer simulation where an input is only known by its distribution e.g., in meteorology, astrophysics, fluid dynamics, biology, operations research, et cetera.
This complex simulation could be a Monte Carlo simulation, a finite-element or finite-volume simulation, or a stochastic model. 
Within reasonable budget, integration using Monte Carlo may not be feasible and alternative numerical integration schemes are needed that require fewer function evaluations.

Bayesian quadrature (\bq)---a means of constructing posterior measures over the unknown value of the integral \citep{OHagan1991, Diaconis1988,Briol2019}---mitigates a high sample demand by encoding known or assumed structure of the integrand such as smoothness or regularity, usually via a Gaussian process (\gp).
With its increased `data'\footnote{See \eg \citet{Hennig2015} and \citet{Cockayne2017} for a discussion on `data' in numerical solvers.} efficiency, \bq~is a natural choice when function evaluations are precious \citep{Rasmussen2003}.
In the past, \bq~has been applied in reinforcement learning \citep{Paul2018}, filtering \citep{Kersting2016},
and has been extended to probabilistic integrals \citep{Osborne2012b,Osborne2012a,Gunter2014,Chai2019}.

A complementary approach to sample efficiency is to make use of related, cheaper \emph{secondary} information sources.
The task of finding approximations to computationally demanding numerical models is an area of active research all on its own (see \eg \citealp{Benner2017}).
Secondary information sources of reduced cost and quality include numerical models that are run at a lower resolution (\eg a coarser grid in a fluid dynamics application), model simplifications by neglecting details or by using an approximate model that is easier to solve numerically, and analytic approximations.
A primary source could be an elaborate Earth system model to simulate anthropogenic climate change. There exist a plethora of such models and secondary sources might parametrize important effects like albedo or neglect detailed land surface processes or ocean biogeochemistry \citep{IPCC2013}.

Multi-source modeling is a statistical technique for harvesting information from related functions by constructing correlated surrogates over multiple sources.
When the information sources are hierarchical in that they are ordered from most to least informative, this concept is known as multi-fidelity modeling \citep{Kennedy2000,Peherstorfer2018,Forrester2007,LeGratiet2014}.
The notion of \emph{multi-source} models is more overarching and includes settings in which sources do not exhibit an easily identifiable order, if any.
Each of the sources has its own \emph{cost} function that quantifies the cost of evaluating the source at a certain input.
An input-dependent cost might arise when the simulation run to query the integrand needs to be refined for certain values of the input to ensure numerical stability. 
A linear instance of a multi-source model is a multi-output \gp~aka. co-kriging \citep{Alvarez2011a}.
\bq~with multi-output \gp s to integrate several related functions has been studied by \citet{Xi2018}, who impose properties on data---which they assume given---to prove theoretical guarantees and consistency of the Bayes estimator.

\bq~leverages active learning schemes similar to Bayesian optimization \citep{Shahriari2016} or experimental design \citep{Atkinson2007,Yu2006}.
Through its argmax, an acquisition function identifies optimal future locations to query the integrand according to a user-defined metric.
Metrics of interest in \bq~are information gain on the value of the integral or its predictive variance.
By optimizing the target \emph{per cost}, active multi-source \bq~(\amsbq) trades off improvement on the target (the integral) and ressources spent.
In Bayesian optimization, this setting has been explored by \citet{Poloczek2017}.

We summarize our contributions:
\begin{itemize}[leftmargin=*, topsep=0pt, itemsep=0pt]
\item We lay the foundations for active \bq~for the task of integrating an expensive function that comes with cheaper approximations. 
We assign cost to function evaluations and generalize \vbq~acquisition functions to acquisition \emph{rates} that trade off improvement on the integral against cost.
\item We find that some rates induce sane, others pathological acquisition policies. Pathologies were not present in the common \vbq~acquisition schemes that all give rise to the same degenerate policy, regardless of the acquisition's \emph{value}. Cost-adapted rate policies do depend on these values and are thus intricately tied to the meaning of the acquisition function that encodes progress on the quadrature task. Simply put, \emph{all} considered (even pathological) multi-source acquisition policies collapse onto a single policy for \vbq, as a corner case of \amsbq.
\item We conduct proof-of-concept experiments which show that \amsbq~improves upon \vbq~in that it spends less budget on learning the integral to an acceptable precision.
\end{itemize}

\section{MODEL}
\label{sec:model}

We wish to estimate the integral over the information source of interest (the \textit{primary} source), w.l.o.g. indexed by 1, $f_1:\Omega\mapsto \Re$, $\x\mapsto f_1(\x)$ and integrated against the probability measure $\pi$ on $\Omega\subseteq\Re^D$, 
\begin{equation}
\label{eq:f1int}
  \fint =: \int_\Omega f_1 (\x) \dd \pi(\x)
\end{equation}
in presence of $L-1$ not necessarily ordered or orderable \textit{secondary} information sources $f_2, \dots, f_L$, with $f_l:\Omega\mapsto \Re$.
Each source $l\in\L = \{1,\dots, L\}$ comes with an input-dependent cost $c_l(\x)$ which must be invested to query $f_l$ at location $\x$.
For ease of interpretation and numerical stability we set $c:\L\times \Omega\mapsto [\delta, 1]$ and $0<\delta\leq 1$.
This is equivalent to assuming there exists a $c_{\text{min}}>0$ and a $c_{\text{max}}<\infty$ s.t. $c_{\text{min}} \leq c_l(\x) \leq c_{\text{max}}$ and then normalizing w.r.t. $c_{\text{max}}$, \ie $\delta = \frac{c_{\text{min}}}{c_{\text{max}}}$.
In other words, no query takes an infinite amount of resources, nor does any evaluation come for free.
Normalization is not required and in practice, neither $c_{\text{max}}$ nor $c_{\text{min}}$ need to be known.

In this section we review the tools for building a statistical model that allows us to harvest information from both the primary and the secondary sources for learning the integral $\fint$ of Eq.~(\ref{eq:f1int}), before turning to the decision theoretic problem of how to actively select locations and sources to query next in Section~\ref{sec:acq}.

\subsection{VANILLA BQ}
\label{sec:vbq}

Let $f:\Omega \mapsto \Re$, $\x\mapsto f(\x)$ be a function and $\pi$ a probability measure on $\Omega\subseteq\Re^D$ that has an intractable integral $\intangle{f} = \int_\Omega f(\x) \dd \pi(\x)$.
In vanilla Bayesian quadrature (\vbq), we express our epistemic uncertainty about the value of $\intangle{f}$ through a random variable $\rz$.
The distribution over $\rz$ is obtained by integrating a Gaussian process (\gp) prior that is placed over the integrand $f$, i.e. $f \sim \GP(m, k)$, where $m: \Omega\mapsto\Re$, $\x\mapsto m(\x)$ denotes the prior mean function and $k: \Omega\times\Omega\mapsto\Re$, $(\x, \x')\mapsto k(\x, \x')$ the covariance function or kernel. 
Observations come in form of potentially noisy function evaluations\footnotemark~$y = f(\x) + \eps$ with $\eps\sim \N (0, \sigma^2)$.
\footnotetext{Noise free evaluations are usually assumed in \bq, but this might not be true for a black-box integrand.}
Let $\sX$ denote the matrix of $N$ input locations $\sX=[\x_1\ \dots\ \x_N]\Trans$ and $\y=f(\sX) + \boldsymbol{\eps}$ the set of corresponding observations, summarized in $\Dcal=\{\sX, \y\}$ (see \citealp{Rasmussen2006} for an introduction to \gp~inference).  
With the closure property of \gp s, the posterior over $\rz$ when conditioning on $\Dcal$ is a univariate Gaussian distribution with posterior mean $\Exp [\rz\g\Dcal] = \intangle{m_{\Dcal}}$ and variance $\V [\rz\g\Dcal] = \int_\Omega \int_\Omega k_{\Dcal} \dd \pi(\x) \dd \pi(\x') \eqqcolon \iintangle{k_{\Dcal}}$ that are integrals over the \gp's posterior mean $m_\Dcal (\x)$ and covariance $k_\Dcal (\x, \x')$. 
These expressions are detailed below for the general multi-source case that \vbq~is a subset of and further derivations can be found in \citet{Briol2019}.
So as not to replace an intractable integral by another intractable integral, the kernel $k(\x,\x')$ is chosen to be integrable against $\pi(\x)$.

\subsection{MULTI-SOURCE MODELS}
\label{sec:multi-source-models}

We consider linear multi-source models, which can equally be phrased as multi-output Gaussian processes \citep{Alvarez2011a} over the vector-valued function $\fvec = [f_1, \dots, f_L]$, $\fvec: \Omega \mapsto \Re^L$. 
Non-linear models for multi-source modeling exist and have been considered by \citet{Perdikaris2017}. 
They do however come with the additional technical difficulty that the model may not be integrable analytically---a sensible pre-requisite for \bq---and are thus another beast altogether.
The notation mimics the single-output case, that is, $\fvec \sim \GP(\m, \sK)$, where $\sK$ is an $L\times L$ matrix-valued covariance function.
More precisely, the covariance between two sources $f_l$ and $f_{l'}$ at inputs $\x$ and $\x'$ is $\cov [f_l(\x), f_{l'} (\x')] = k_{ll'} (\x, \x')$.
The kernel $k_{ll'}(\x,\x')$ encodes not only characteristics of the individual sources (\eg smoothness), but crucially the correlation between them. 
In the multi-source setting, observations come in source-location-evaluation triplets $(l, \x, y_l)$ with $y_l = f_l (\x) + \eps_l$ and source-dependent observation noise $\eps_l \sim \N(0,\sigma_l^2)$ as usually only one element of $\fvec$ is being observed (see supplementary material (\suppmat) for alternative representation as linear observations).

The dataset $\Dcal = \{\ldat, \sX, \y_{\ldat}\}$ contains $N$ data triplets from evaluating elements of $\fvec$ at $N$ locations $\sX=[\x_1\ \dots\ \x_N]\Trans$ with corresponding sources $\ldat=[l_1\ \dots\ l_N]\Trans$ and observations $\y_{\ldat} =  [f_{l_1}(\x_1) + \eps_{l_1}\ \dots\ f_{l_N}(\x_N) + \eps_{l_N}]\Trans$. 
The \gp~posterior over $\fvec$ has mean and covariance
\begin{equation}
\label{eqn:posterior-multi-source}
\begin{aligned}
  m_{l | \Dcal} (\x) &= m_l(\x) + \kvec_{l\ldat} (\x, \sX) \sG_{\ldat}(\sX)\inv (\y_{\ldat} - \m_{\ldat}(\sX)),\\
  k_{l l' | \Dcal} (\x, \x') &= k_{ll'} (\x,\x') - \kvec_{l\ldat} (\x, \sX) \sG_{\ldat}(\sX)\inv \kvec_{\ldat l} (\sX, \x'),
\end{aligned}
\end{equation}
with the kernel Gram matrix $\sG_{\ldat}(\sX) = \sK_{\ldat\ldat}(\sX,\sX) + \sSigma_{\ldat}\in\Re^{N\times N}$ and $\sSigma_{\ldat} = \diag (\sigma_{l_1}^2, \dots, \sigma_{l_N}^2)$.
A summary of the notation used can be found in Table~\ref{tbl:notation} in the supplementary material.

\subsection{MULTI-SOURCE BQ}
\label{sec:multi-source-quad}

The multi-source model of Section~\ref{sec:multi-source-models} can be integrated and gives rise to a quadrature rule similar to \vbq~(cf. sec. \ref{sec:vbq}).
Let $\rz$ denote the random variable representing the integral of interest $\fint$ of Eq.~(\ref{eq:f1int}).
The posterior over $\rz$ given data triplets $\Dcal$ is a univariate Gaussian with mean and variance
\begin{equation}
\begin{aligned}
  \Exp [Z \g \Dcal] &= \intangle{m_1} + \intangle{\kvec_{1 \ldat} (\cdot,\sX)} \sG_{\ldat}(\sX)\inv (\y_{\ldat} - \m_{\ldat}(\sX)),\\
  \V [Z \g \Dcal]  &= \iintangle{k_{11}} - \intangle{\kvec_{1\ldat} (\cdot,\sX)} \sG_{\ldat}(\sX)\inv \intangle{\kvec_{\ldat 1} (\sX, \cdot)},
\end{aligned}
\end{equation}
where $\intangle{\kvec_{1 \ldat} (\cdot,\sX)} = \int_\Omega \kvec_{1 \ldat} (\x, \sX) \dd \pi(\x)$ is the kernel mean and $\iintangle{k_{11}}=\int_{\Omega}\int_{\Omega} k_{11} (\x, \x') \dd \pi(\x) \dd \pi(\x')$ the initial error, both of source 1.
Just as in \vbq, the kernel is required to be integrable analytically.

We choose an intrinsic coregionalization model (\icm) \citep{Alvarez2011a} with kernel
\begin{equation}
  \label{eq:coregkern}
  k_{ll'} (\x,\x') = \sB_{ll'} \kappa(\x,\x'),
\end{equation}
where $\sB\in \Re^{L\times L}$ is a positive definite matrix. 
Eq.~(\ref{eq:coregkern}) is a simple extension of a standard kernel $\kappa(\x, \x')$ to the multi-source case which factors the correlation between the sources and input locations. 
If $\kappa(\x, \x')$ is integrable analytically, $k_{ll'} (\x,\x')$ will be, too, and thus retains the favorable property of a \bq-kernel. 
A typical choice for $\kappa$ is the squared-exponential, aka. \textsc{rbf} kernel $\kappa(\x,\x') = \exp (\nicefrac{-\|\x-\x'\|_2^2}{2\lambda^2})$ with no dependence on the sources $l$ and $l'$. 
This model can easily be extended \eg to a linear model of coregionalization (\textsc{lmc}) without challenging integrability of $k$. 
This would untie the lenthscales between sources, but would also introduce $L-1$ additional generally unknown kernel parameters.
The simpler \textsc{icm} is also used by \citet{Xi2018} to establish convergence rates for a multi-output \bq~rule.

\section{ACTIVE LEARNING}
\label{sec:acq}

\begin{figure*}[thb]
  \begin{center}
    \includegraphics{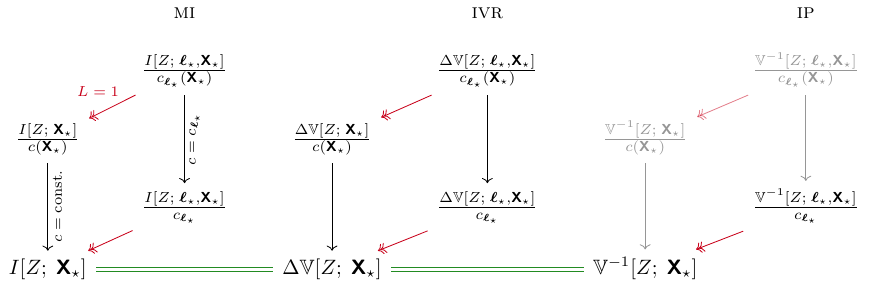}%
  \caption{
  The multi-source acquisition-cube for a few of the possible acquisition functions. \mi, \ivr, and \ip~stand for `mutual information', `integral variance reduction', and `integral precision', respectively. %
  The forward arrows ({\color{mred}\protect\rotatebox[origin=c]{-150}{$\twoheadrightarrow$}}) denote the special case of one source only ($L=1$) as in the case of \vbq.
  The downward facing arrows ($\downarrow$) denote the special case where the cost $c$ is not dependent on the locations $\sX_\star$. 
  The double-lines ({\color{forestgreen}$\xlongequal{~~~}$}) between nodes denote that these acquisition functions are equivalent in the sense that they yield the same optimal $\sX_{\star}$.
  The two grayed-out acquisitions for \ip~highlight that they exhibit non-favorable behavior (cf.~Section~\protect\ref{sec:multi-source-acq}).
  The bottom front row in the cube denotes the special case of \vbq~($L=1$ and $c(\sX_\star)=\text{const.}$) where all three acquisition policies (\mi, \ivr, \ip) coincide.
  }
  \label{fig:cube}
  \end{center}
\end{figure*}

\emph{Active learning} describes the automation of the decision-making about prospective actions to be taken by an algorithm in order to achieve a certain learning objective.
A heuristic measure of improvement towards the specified goal (here: learning the value of an integral) is defined through a \emph{utility function}.
It transfers the decision-theoretic problem to an optimization problem, but usually an unfeasible one.
Therefore, the utility is commonly approximated by an \emph{acquisition function}.
Optimizing the acquisition function induces an \emph{acquisition policy} that pins down what action to take next.
To obtain a sequence of actions, the considered method (here: \bq) is placed in a loop where it is iteratively fed with $N_\star$ new observations $(\ldat_\star, \sX_\star, \y_{\ldat_\star})$ in the general multi-step look-ahead (\emph{non-myopic}) approach.
A \emph{myopic} approximation is to instead optimize for a single new observation triplet $(l_\star, \x_\star, y_{l_\star})$ at a time.
Besides feasibility, the lack of exact model knowledge motivates a loop in which the model is repeatedly updated with new observations.

In Section~\ref{sec:nocost_acq}, we recapitulate several utilities that are commonly used for \vbq.
All these utilities give rise to the same acquisition policy in the absence of cost and are thus not greatly differentiated between in the literature.
Intriguingly, the policies do not coincide for \amsbq~if cost is accounted for in the acquisition functions, as will be shown and discussed in Section~\ref{sec:multi-source-acq}.

\subsection{MULTI-SOURCE BQ ACQUISITIONS}

\label{sec:nocost_acq}
In the absence of any notion of evaluation cost (or if all sources come at the same cost), the utility functions from \vbq~generalize straightforwardly to the multi-source case.
The \vbq~case can be recovered by setting the number of sources to one.

\subsubsection{Mutual Information}
\label{sec:mutual-information}

From an information theoretic perspective, new source-location pairs $(\ldat_\star, \sX_\star)$ can be chosen such that they jointly maximize the mutual information (\mi) $I [\rz; \y_{\ldat_\star}]$ between the integral $\rz$ and a set of new but yet unobserved observations $\y_{\ldat_\star}$ with $y_{l_\star^i} = f_{l_\star^i} (\x_\star^i) + \epsilon_{l^i_\star}$.
In terms of the individual and joint differential entropies over $\rz$ and $\y_{\ldat_\star}$, $I [\rz; \y_{\ldat_\star}] = H[\rz] + H[\y_{\ldat_\star}] - H[\rz, \y_{\ldat_\star}]$.
Sections~\ref{sec:multi-source-models} and \ref{sec:multi-source-quad} imply that both $\rz$ and $\y_{\ldat_\star}$ are normally distributed and so is their joint.
The differential entropy of a multivariate normal distribution with covariance matrix $\sA \in \Re^{M\times M}$ is $H = \frac{M}{2} \log (2\pi e) + \frac{1}{2} \log |\sA|$.
Since there is no explicit dependence on the value of $\y_{\ldat_\star}$, we (sloppily) express the mutual information as a function of the new source-location pairs $(\ldat_\star, \sX_\star)$,
\begin{equation}
\label{eqn:mutinfo}
  I [\rz; \ldat_\star, \sX_\star]= - \frac{1}{2} \log\left(1 - \rho^2_{1 \ldat_\star | \Dcal} (\sX_\star) \right),
\end{equation}
where we introduce the scalar correlation
\begin{equation}
\label{eqn:rho2}
  \rho^2_{1 \ldat_\star | \Dcal} (\sX_\star) := \frac{ \intangle{\kvec_{1 \ldat_\star | \Dcal} (\cdot, \sX_\star)}\, \sV_{\ldat_\star | \Dcal}\inv (\sX_\star)\, \intangle{\kvec_{\ldat_\star\! 1 | \Dcal} (\sX_\star, \cdot)} }{\V[\rz \g \Dcal]},
\end{equation}
$\in [0,1]$, with the noise-corrected posterior covariance matrix $\sV_{\ldat_\star | \Dcal} (\sX_\star) = \sK_{\ldat_\star\ldat_\star | \Dcal}(\sX_\star, \sX_\star) + \sSigma_{\ldat_\star}\in \Re^{N_{\star}\times N_{\star}}$.
In the one-step look-ahead case ($N_{\star}=1$), 
\begin{equation}
  \rho_{1 l_\star | \Dcal} (\x_\star) = \frac{\intangle{k_{1 l_\star | \Dcal} (\cdot, \x_\star)}}{\sqrt{v_{l_\star | \Dcal} (\x_\star)\, \V [\rz \g \Dcal]}}
\end{equation}
is the bivariate correlation between $\rz$ and $y_{l_\star}$.

\subsubsection{Variance-Based Acquisitions}
\label{sec:intvarred}
Variance-based approaches attempt to select $(\ldat_\star, \sX_\star)$ such that the variance on $\rz$ shrinks maximally. 
As \mi, the integral variance reduction (\ivr) normalized by the current integral variance $\V[\rz \g \Dcal]$ can be written in terms of correlation $\rho$ as
\begin{equation}
\label{eqn:ivr}
\begin{aligned}
  \frac{\Delta \V[\rz; \ldat_\star, \sX_\star]}{\V[\rz \g \Dcal]} &= \frac{\V[\rz \g \Dcal] - \V [\rz \g \Dcal \cup (\ldat_\star, \sX_\star, \y_{\ldat_\star})]}{\V[\rz \g \Dcal]}\\
  & = \rho^2_{1 \ldat_\star  | \Dcal} (\sX_\star).
\end{aligned}
\end{equation}
Eq.~\eqref{eqn:ivr} is a monotonic transformation of Eq.~\eqref{eqn:mutinfo} and therefore, both utility functions share the same global maximizer $\sX_{\star}$.
In fact, any monotonic transformation of Eq.~\eqref{eqn:rho2}, whether interpretable or not, gives rise to the same acquisition policy.
This is because the policy only depends on the \emph{locations}, but not the \emph{value} of the utility function's global maximum.
Hence, in \vbq~it is equivalent to consider maximal shrinkage of the variance of the integral, minimization of the integral's standard deviation, or maximal increase of the integral's precision (\ip), to name a few---they all lead to the same active learning scheme and have thus not been greatly distinguished between in previous work on active \vbq.

\subsection{COST-SENSITIVE ACQUISITIONS}
\label{sec:multi-source-acq}

When there is a location and/or source-dependent cost associated to evaluating the information sources (cf.~Section~\ref{sec:model}), the utility function should trade off the improvement made on the integral against the budget spent for function evaluations.
This is achieved by considering the ratio of a cost-insensitive \bq~utility and the cost function $c_{\ldat_\star} (\sX_\star) = \sum_{i=1}^{N_\star} c_{l_i} (\x_i)$.
Such a ratio can be interpreted as an acquisition \emph{rate} and bears the units of the utility function divided by units of cost.
The notion of a rate becomes clearer when considering for example the mutual information utility Eq.~\eqref{eqn:mutinfo} with cost measured in terms of evaluation time: the unit is $\frac{\text{bits}}{\text{second}}$, \ie a rate of information gain.\\
This construction has an important consequence: Modification of the \vbq~utility function (\ie the numerator), even by a monotonic transformation, changes the maximizer of the cost-adapted acquisition rate and hence, also the acquisition \emph{policy}.
In other words, the degeneracy of \bq~acquisition functions in terms of the policy they induce in the absence of cost is lifted when evaluation cost is included, firstly, because the argmax of each acquisition is shifted differently with cost, and, secondly, because acquisition \emph{values} from different sources are discriminated against each other now.
As will be discussed below, not all monotonic transformations yield a sensible acquisition policy; indeed, some display pathological behavior.

The adapted non-myopic acquisition rates for the \bq~utilities mutual information (\mi~Eq.~\eqref{eqn:mutinfo}) and integral variance reduction (\ivr~Eq.~\eqref{eqn:ivr}) are 
\begin{align}
\label{eqn:mi_nonmyopic}
\alpha_{\ldat_\star}^{\mi} (\sX_\star) &\coloneqq \frac{-\log\left(1 - \rho^2_{1 \ldat_\star | \Dcal} (\sX_\star) \right)}{c_{\ldat_\star} (\sX_\star)}\\
\label{eqn:vr_nonmyopic}
  \alpha_{\ldat_\star}^{\ivr} (\sX_\star) &\coloneqq \frac{\rho^2_{1 \ldat_\star | \Dcal} (\sX_\star)}
      {c_{\ldat_\star}(\sX_\star)},
\end{align}
where we have dropped the factor $\nicefrac{1}{2}$ in \mi~as an arbitrary scaling factor.
It is evident that these acquisition rates do no longer share their maximizer; yet they still induce a meaningful acquisition scheme.
Both \mi~and \ivr~have the property to be zero at $\rho^2=0$ and thus never select points $\sX_{\star}$ that are uncorrelated with the integral $\rz$, no matter the cost, \eg locations that have already been observed exactly (with $\sigma^2=0$).
Such points do not update the posterior of the integral $\rz$ when conditioned on.
In \vbq~these locations are the minimizers of all acquisition functions and thus excluded no matter their value.
This is not ensured for the cost-adapted acquisition rates and therefore, they additionally require the numerator to be zero at $\rho^2=0$.
Hence, not every monotonic transformation of the \bq~utility produces a sane acquisition policy in the presence of cost.\\
Consider for example the valid transformation $\rho^2 \mapsto \rho^2 - 1$, which is $-1$ at $\rho=0$.
Maximizing this utility function corresponds to maximizing the negative integral variance, \ie minimizing the integral variance, which is very commonly done in \vbq.
Since $\rho^2\in [0,1]$, $\rho^2 -1$ is negative everywhere and gets larger (takes a smaller negative value) with larger cost.
Hence when maximized, this acquisition would favor expensive evaluations.\\
More subtle is the misbehavior of the integral precision (\ip) which is positive everywhere and has the desired behavior of favoring low-cost evaluations.
In terms of the squared correlation $\rho^2\in[0, 1]$ from Eq.~\eqref{eqn:rho2} (with simplified notation for convenience), the numerator of the \ip~acquisition rate can be written as $(1 - \rho^2)\inv$.
This expression is non-zero at $\rho^2 = 0$ and therefore, it does not exclude points of zero correlation when they come at sufficiently cheap cost, and in experiments we observe it getting stuck re-evaluating at the location of minimum cost ad infinitum.
We conjecture that this is because \ip~only encodes an absolute scale of the integral variance but does not quantify any ``improvement'' on the integral value.
\begin{figure}[!t]
\begin{center}
    \includegraphics{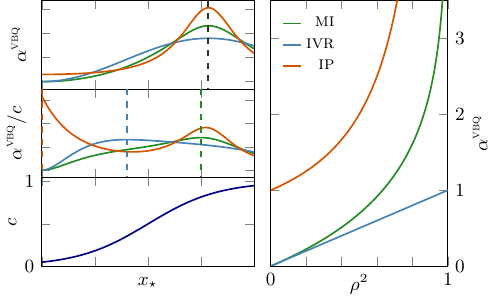}%
\caption{\textit{right:} \vbq~acquisitions $\alpha^{\vbq}$ as functions of the squared correlation $\rho^2 (\sX_\star)$; \textit{left:} \vbq~acquisitions as a function of univariate $x_\star$ and  myopic step ($N_{\star}=1$) for a synthetic $\rho^2(x_\star)$. Without cost, their maximizers coincide (top), but when divided by an input-dependent cost $c(x_\star)$ (bottom), the maximizers disperse (indicated by the dashed vertical lines) (middle). For implications, cf. Section~\ref{sec:multi-source-acq}.}
\label{fig:intuition}
\end{center}
\end{figure}
\\
Figure~\ref{fig:cube} illustrates the augmentation of utility functions from \vbq~with multiple information sources and cost.
Figure~\ref{fig:intuition} displays the behavior of a few acquisitions, \mi, \ivr, and \ip.
The right plot shows these acquisitions as used in \vbq~in terms of the squared correlation $\rho^2\in[0, 1]$ (Eq.~\eqref{eqn:rho2}) in the absence of cost.
All acquisitions are strictly monotonically increasing functions of $\rho^2$.
Among the sane acquisition rates that are zero at $\rho^2=0$, the differences in the corresponding policy can also be understood from the functional dependence on $\rho$.
\mi~diverges at perfect correlation $\rho^2\rightarrow 1$.
Therefore, and since the cost $c$ lies in $[\delta, 1]$, \mi~will always take a `perfect step' to learn the integral exactly, \ie it will always select the points $\sX_\star$ with correlation $\rho^2 (\sX_\star) = 1$, if the step is available and no matter the cost.
\ivr, however, is finite at $\rho^2=1$ and trades off cost against correlation even if the perfect $\sX_\star$ with $\rho^2(\sX_\star)=1$ exists.
These interpretations are reinforced by the left three plots of Figure~\ref{fig:intuition}, in which we plot \mi, \ivr, and \ip~versus a univariate $x_{\star}$ for the synthetic choice $\rho^2(x_\star)=0.95\sin^2(10 x_{\star})$, $x_{\star}\in[0, 0.2]$ and a myopic step ($N_{\star}=1$).
In the pure \vbq~situation, the locations of all their maxima coincide, but as soon as a non-constant cost $c(x_\star)$ is applied, the shapes of the acquisition functions become relevant which discriminates their $\sX_\star$ and lifts the degeneracy in policies.
\mi~tends more towards higher correlation than \ivr, the maximizer of which moves further towards locations of lower cost.
While \mi~and \ivr~act differently, they are both sensible choices for acquisition functions in \amsbq. 
In fact for low to mid-ranged values of $\rho^2 \lsim 0.5$ where \mi~is approximately a linear function of $\rho^2$ they roughly coincide.\\
The choice of acquisition ultimately depends on the application and the user, who may choose which measures of improvement on the integral and cost to trade off.
%

\section{EXPERIMENTS}
\label{sec:experiments}

\begin{figure*}[thb]
\begin{center}
    \includegraphics{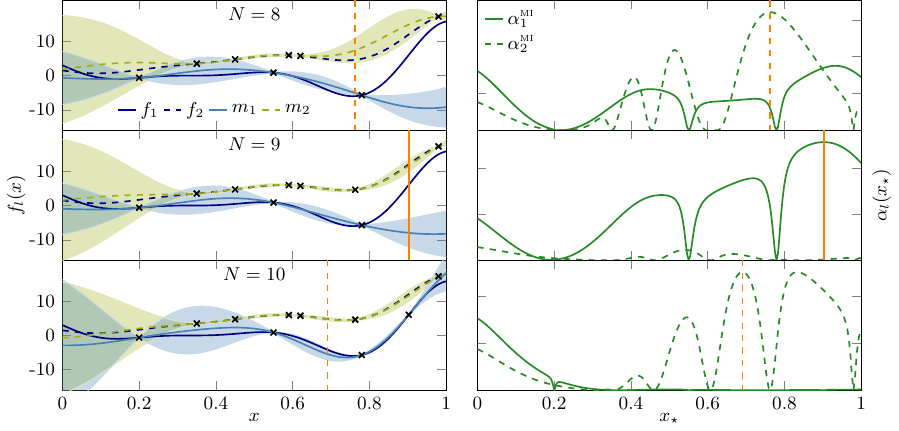}%
\caption{Demonstration of the sequential selection of new source-location pairs to query $\fvec$ using the \mi~acquisition in a two-source setting with source and location dependent cost (Figure~\ref{fig:cost-demo}). \emph{Left column:} The \gp; \emph{right column:} the acquisition function for the primary (solid) and secondary source (dashed) for three consecutive iterations. Vertical orange lines indicate the location and source of the new query.}
\label{fig:demo}
\end{center}
\end{figure*}

The key practical applications for \amsbq~is solving integrals of expensive-to-evaluate black-box functions that are accompanied by cheaper approximations, potentially in a setting where a finite budget is available.
Typical applications are models of complex nonlinear systems that need to be tackled computationally.
With evaluations being precious, the goal is to get a decent estimate of the integral with as little budget as possible, rather than caring about floating-point precision.
In the experiments, we focus on the rear vertices of the acquisition cube Figure~\ref{fig:cube}, \ie multiple sources with source and input-dependent or only source-dependent cost, and separate them into two main experiments:
\begin{enumerate}[leftmargin=*, topsep=0pt, itemsep=0pt]
  \item A synthetic multi-source setting with cost that varies in source and location for the purpose of exploring and demonstrating the behavior of the acquisition functions derived in Section~\ref{sec:acq}.%
  \item An epidemiological model of the spread of a disease with uncertain input, in which two sources correspond to simulations that differ in cost as well as quality of the quantity of interest.%
\end{enumerate}
Additionally, we present a bivariate experiment with three sources in the \suppmat~Section~\ref{sec:bivar-line-comb}.
We take a myopic approach to all scenarios in that we optimize the acquisition for a single source-location pair a time.
The implementation of the \gp-model uses \texttt{GPy} \citep{gpy2014} in \texttt{Python 3.7}.

\subsection{MULTI-SOURCE, VARIABLE COST}
\label{sec:multiple-sources-non}

We initially consider a synthetic two-source setting with univariate input.
The cost functions depend on both source and location.
The experiment's purpose is to demonstrate our findings from Section~\ref{sec:acq} and convey intuition about the behavior of the novel acquisition functions.
The sources we consider have been suggested by \citet{Forrester2007} with the primary source $f_1(x) = (6x - 2)^2  \sin(12x - 4)$ and the secondary source $f_2(x) = \frac{1}{2} f_1(x) + 10x$ for $x\in[0,1]$.
The cost functions both take the form of a scaled and shifted logistic function in a way that the cost lies in $(0,1]$ (cf. Figure~\ref{fig:cost-demo} in the \suppmat).
The costs of both sources converge to the same value close to $x_\star=0$; for larger $x_\star$, $f_2$ is two orders of magnitude cheaper than $f_1$.
Figure~\ref{fig:demo} shows snapshots of three consecutive query decisions taken by the \mi~multi-source acquisition.
The \gp~model (depicted in the left column) has been initialized with 3 datapoints in the primary and 5 in the secondary source and merely the noise variance was constrained to $10^{-2}$.
The \mi~acquisition given the current state of the \gp~is shown on the right---the top left frame is shown for \mi, \ivr, and \ip~in Figure~\ref{fig:forrester_all_acq} in the \suppmat~to emphasize the pathology of \ip~and to highlight the subtle difference between \mi~and \ivr~in practice.
The acquisition function is optimized using the \textsc{l-bfgs-b} optimizer in \texttt{scipy}.
We observe that \amsbq~does not query $f_2$ where the source costs are almost identical for $x_\star \lsim 0.2$ (see Figure~\ref{fig:finalforrester} in supp. mat.).
This is because the two sources are not perfectly correlated and evaluating $f_1$ always conveys more information about $\rz$ than $f_2$.
The fact that $c_2$ decreases with increasing $x_\star$ is nicely represented in the increasing height of the maxima of the dashed acquisition function for the secondary source in the top left frame of Figure~\ref{fig:demo}.

\begin{figure}[thb]
\begin{center}
    \includegraphics{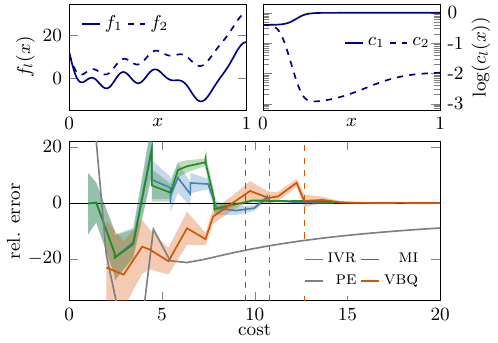}%
\caption{\textit{Top left:} the wigglified Forrester functions with $f_1$ and $f_2$ primary/secondary source, respectively; \textit{top right:} the cost functions used; \textit{bottom:} relative error $\nicefrac{\Exp[\rz]-\fint}{\fint}$ with two std.~deviations (shaded) as a function of normalized cost for the \amsbq~acquisitions \mi~and \ivr~compared to \vbq~and a percentile estimator (\textsc{pe}). Vertical dashed lines are a visual help to indicate the cost spent to achieve acceptable accuracy.}
\label{fig:wiggly}
\end{center}
\end{figure}

For assessing the performance of \amsbq, we compare against \vbq~and a percentile estimator (\textsc{pe}) that both operate on the primary source.
The latter is obtained by separating the domain into intervals that contain the same probability mass and summing up the function values at these nodes.
For the uniform integration measure used here, this is equivalent to a right Riemann sum.
We assume that \gp~inference comes at negligible cost as compared to the function evaluations and thus consider cost to be incurred purely by querying the information sources.\\
To render the integration problem more difficult, we modify the Forrester functions to vary on a smaller length scale by adding a sinusoidal term and adapting some parameters, s.t.
$f_1(x) = (6x - 2)^2  \sin(12x - 4)  - (2-x)^2 \sin(36x)$ and $f_2(x) = \frac{3}{4} f_1(x) + 16\left(x - \frac{1}{2}\right) + 10$ which we integrate from 0 to 1 against a uniform measure (cf. Figure~\ref{fig:wiggly}, top left).
To avoid over- or underfitting, we set a conservative gamma prior on the lengthscale with a mode at a small fraction of the domain $[0,1]$ for both \vbq~and \amsbq, and assume zero observation noise.
With six\footnote{Due to the construction of $\sB = \sW\sW\Trans + \diag (\vec{\eta})$} more hyperparameters than \vbq, \amsbq~is more prone to over-/underfitting, and we further set a prior on the coregionalization matrix $\sB$ (cf. Section~\ref{sec:multi-source-quad}) with parameters estimated from the initial three data points using empirical Bayes.
This is to avoid initial over- or under-estimation of the correlation between sources, which would either cause the active scheme to select only $f_2$ or only $f_1$, respectively.
Compared to the previous experiment, the cost is changed to have a minimum, but still composed of a sum of logistic functions and normalized to be in $(0,1]$ (Figure~\ref{fig:wiggly}, top right).
The effect of these cost functions on the final state is depicted in the \suppmat, Figure~\ref{fig:finalwiggly}.
Furthermore, this setting reveals the pathology of the \ip~acquisition (cf.~Section~\ref{sec:multi-source-acq}) that everlastingly re-evaluates the secondary source at the location of minimal cost.
The convergence behavior of the well-behaved acquisition functions \mi~and \ivr~are displayed in Figure~\ref{fig:wiggly} (bottom) in comparison to \vbq~and \textsc{pe}.
The hyperparameters of the \gp~are optimized after every newly acquired node, both for \vbq~and \amsbq.
Figure~\ref{fig:wiggly} shows the superior performance of both \amsbq~methods in arriving close the true integral with little budget.
The vertical jumps in the \amsbq~methods occur when $f_2$ is evaluated at cheaper cost.

\subsection{A SIMULATION OF INFECTIONS}
\label{sec:multi-fidel-simul}

\begin{figure*}[thb]
\begin{center}
    \includegraphics{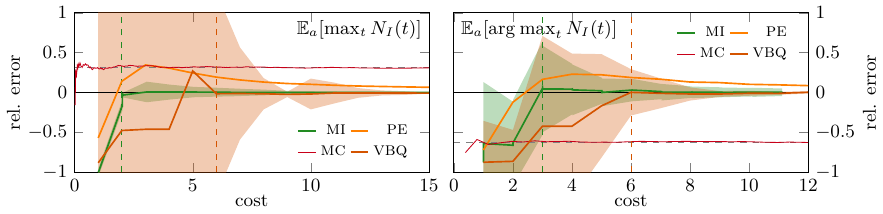}%
\caption{Relative error vs. budget spent for the \textsc{sir} model for the max number of simultaneously infected individuals (left) and for the time after which the maximum occurs (right). Primary source has cost 1.
}
\label{fig:sir}
\end{center}
\end{figure*}

We now consider multi-source models in which sources come with input-independent cost, a.k.a. multi-fidelity models (bottom rear \mi~vertex in Figure~\ref{fig:cube}).
We choose an epidemiological model in which evaluating the primary source requires running numerous stochastic simulations and the secondary source solves a system of ordinary differential equations.
Epidemiological models deal with simulating the propagation of an infectious disease through a population.
The \textsc{sir} model forms the base for many compartmental models and assumes a population of fixed size $N$ where at any point in time, each individuum is in one of three states---susceptible, infected, and recovered (\textsc{sir})---with sizes $N_S, N_I$, and $N_R$ \citep{Kermack1927}.
The dynamics are determined by stochastic discrete-time events of individuals changing infection state, for which Poisson processes (\ie exponentially distributed inter-event times) are commonly assumed (see \eg \citealp{Daley1999}).
In the thermodynamic limit where $N$ is large, the average dynamics is governed by a system of \textsc{ode}s that does not admit a generic analytic solution.
There are two parameters in the \textsc{sir} model: the infection rate $a$, and the recovery rate $b$.
Model details and experiment setup can be found in Section~\ref{sec:sirdetails} (\suppmat).

For the \amsbq~experiment, we assume that we know $b$, but we are uncertain about $a$. 
We are interested in the expected maximum number of simultaneously infected individuals $\Exp_a[\max_t N_I(t)]$ and the time this maximum occurs $\Exp_a[\argmax_t N_I(t)]$, which might be relevant for vaccination planning.
Querying the primary source $f_1$ for the quantities of interest as a function of $a$ requires numerous realizations of a stochastic four-compartments epidemic model (an extension to the \textsc{sir} model) using the Gillespie algorithm \citep{Gillespie1976,Gillespie1977}.
For each trajectory, the maximum value and time are computed and henceforth averaged over.
In our implementation, each query of $f_1$ takes $\sim 16\,\mathrm{s}$ on a laptop's \textsc{cpu}.
The secondary source $f_2$ solves the system of \ode s for given $a$ and computes the maximum value and time for the resulting function $N_I(t)$, which takes about $8\cdot10^{-3}\mathrm{s}$ to evaluate.
As in previous experiments, we set a gamma prior on the lengthscale, a prior on the coregionalization matrix $\sB$, and the noise variance to zero as in Section~\ref{sec:multiple-sources-non}.
Both \vbq~and \amsbq~are given the same initial value of $f_1$, and \amsbq~additionally gets the value of $f_2$ at the same location, as well as one more random datum from $f_2$.
This is justified since \amsbq~needs to learn more hyperparameters than \vbq~and secondary source evaluations are very cheap.
Otherwise, if the initial evaluations of $f_2$ were further apart than the prior lengthscale from the locations of the initial primary datum, virtually zero correlation would be inferred between the sources, and the primary source would be evaluated until a sampled location roughly coincides with locations where the secondary sources have been evaluated.

\begin{figure}[thb]
\begin{center}
   \includegraphics{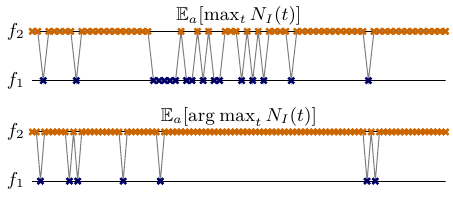}%
\caption{Evaluation sequences of primary and secondary source in the \textsc{sir} experiment.}
\label{fig:evalorder}
\end{center}
\end{figure}

Figure~\ref{fig:sir} shows the relative error of the \amsbq~estimator against normalized cost as compared to \vbq~and \textsc{pe} for $\Exp_a[\max_t N_I(t)]$ (left) and $\Exp_a[\argmax_t N_I(t)]$ (right).
The horizontal dashed line shows $\intangle{f_2}$, \ie the integral of the secondary source with one evolution of a Monte Carlo estimator of $f_2$.
This illustrates that simply using the secondary source for the integral estimate might be computationally cheap, but results in an unknown bias.
In the left plot, \amsbq~achieves a good estimate with one additional evaluation of $f_1$ only, while \vbq~takes another six evaluations.
Again, the vertical jumps for \amsbq~are caused by evaluations of $f_2$.
The initial high confidence on the integral is caused by the choice of prior on the output scale from the initial data, which is located in the tail of the gamma prior on $a$.
Figure~\ref{fig:evalorder} displays the order in which \amsbq~evaluates primary and secondary source.

\section{DISCUSSION}
\label{sec:discussion}

The multi-source model presented in Section~\ref{sec:multi-source-models} can be extended in various ways to increase its expressiveness by using a more elaborate kernel (\eg one lengthscale per source), or by encoding knowledge about the functions to be integrated, \eg a probabilistic integrand. 
Other applications might come with the complication that the cost function $c$ is unknown a priori and needs to be learned during the active \bq-loop from measurements of the amount of resource required during the queries. 
A simple example was presented in Section~\ref{sec:multi-fidel-simul} where the cost was parameterized by a constant, 
estimated during the initial observations. 
A probabilistic (in contrast to parametric) model upon the cost would induce an acquisition function which is not only conditioned on the uncertain model predictions but also on the uncertain cost predictions. 
Furthermore, as in other active learning schemes, non-myopic steps for acquiring multiple observations $\y_{\ldat_\star}$ at once might be beneficial especially when the multi-source model is already known, and does not benefit from being re-fitted to new data;
or when multiple evaluations of sources come at lower cost than evaluating sequentially.
On the experimental side, more elaborate applications of \amsbq~in areas of active research are reserved for future work.

\subsection{CONCLUSION}
\label{sec:conclusion}

We have placed multi-source \bq~in a loop and thus enabled active learning to infer the integral of a primary source while including information from cheaper secondary sources.
We discovered that utilities that yield redundant acquisition policies in \vbq~give rise to various policies, some desirable and others pathological, when evaluation cost is accounted for.
Our experiments illustrate that with the sensible acquisition functions, the \amsbq~algorithm allocates budget to information retrieval more efficiently than traditional methods do for solving expensive integrals.

\subsubsection*{Acknowledgements}
We thank Andrei Paleyes for software-related support, as well as Philipp Hennig, Motonobu Kanagawa, and Aaron Klein for useful discussions.
AG acknowledges partial funding by the European Research Council through ERC StG Action 757275 / PANAMA and support by the International Max Planck Research School for Intelligent Systems (IMPRS-IS).


\printbibliography[title={\normalsize References}]

                                                                                             
\appendix
\pagebreak

\twocolumn[
\begin{center}
\thispagestyle{empty}
\hsize\textwidth
  \linewidth\hsize \toptitlebar {\centering
  {\Large\bfseries SUPPLEMENTARY MATERIAL \par}
  \vspace{5pt}
  \large{\textbf{Active Multi-Information Source Bayesian Quadrature}}}
 \bottomtitlebar \vskip 0.2in plus 1fil minus 0.1in
\end{center}
]

\section{MULTI-SOURCE MODELS AND MULTI-OUTPUT GPS}
\label{sec:projections}
We have seen in Section~\ref{sec:multi-source-models} that linear multi-source models can be phrased in terms of multi-output \gp s.
Typically, the goal of multi-output \gp s is to model a vector-valued function and observations come as a vector $\vec{y} = \fvec (\x) + \vec{\eps}$, where $\vec{y}\in\Re^L$.
In multi-source models, we wish to observe only elements of $\fvec$, \ie $y_l = f_l(\x) + \eps_l$.
These observations can be written as projections of the vector-valued observations,
\begin{equation}
 y_l = \h_{l}\Trans \vec{y}
\end{equation}
where $\h_l$ denotes a vector with a 1 in the $l^\mathrm{th}$ coordinate and zero elsewhere.
Let $\sY \in \Re^{NL}$ denote the vector of $N$ stacked vector-valued noisy observations $[\vec{y}_1, \dots, \vec{y}_N]$.
Then the corresponding $N$ observations of elements $\ldat = [l_1\ \dots\ l_N]\Trans$ is
\begin{equation}
 \y_{\ldat} = \begin{bmatrix}
       \h_{l_1}\Trans & \cdots & 0\\
       \vdots & \ddots & \vdots\\
       0 & \cdots & \h_{l_N}\Trans
      \end{bmatrix}
      \sY\ =:\ \sH\Trans \sY,
\end{equation}
where $\sH$ is a sparse $NL\times N$ matrix.
Note the delicate notational difference between the $N$ observations of single elements of $\fvec$, $\y_{\ldat}\in\Re^N$, and a single evaluation of the vector-valued function $\vec{y}\in\Re^L$.
The covariance matrix between all of the observations is
\begin{equation}
\begin{aligned}
 &\cov [\y_{\ldat}, \y_{\ldat}] =\\ &\sH\Trans
 \left(
 \underbrace{
 \begin{bmatrix}
       \sK (\x_1, \x_1) & \cdots & \sK (\x_1, \x_N)\\
       \vdots & \ddots & \vdots\\
       \sK (\x_N, \x_1) & \cdots & \sK (\x_N, \x_N)
  \end{bmatrix}}_{\sK (\sX,\sX)} + \sSigma\otimes\one_{N\times N} \right)
  \sH
\end{aligned}
\end{equation}
where $\sSigma = \diag (\sigma_1^2,\dots,\sigma_L^2) \in \Re^{L\times L}$ and $\one_{N\times N}$ is an $N\times N$ matrix with every element a 1.
Also, $\sK (\sX, \sX) \in \Re^{NL\times NL}$.
With the following mappings, we arrive at the multi-source notation introduced in Section~\ref{sec:model};
\begin{equation}
\label{eqn:mappings}
\begin{aligned}
\sH\Trans \sK_{\sX\sX} \sH &= \sK_{\ldat\ldat}(\sX, \sX);\\
\sH\Trans (\sSigma \otimes\one_{N\times N}) \sH &= \sSigma_{\ldat};\\
\sH \sY &= \y_{\ldat}; \quad \text{etc.}
\end{aligned}
\end{equation}
Hence, the notational detour over vector-valued observations $\sY$ is not required and evaluations of individual sources can be incorporated easily in the multi-source model.
From the mappings Eq.~\eqref{eqn:mappings} follow the posterior mean and covariance of the multi-source model Eq.~\eqref{eqn:posterior-multi-source}.

\section{ADDITIONAL PLOTS FOR SECTION \ref{sec:multiple-sources-non}}

Section~\ref{sec:multiple-sources-non} showed two examples to demonstrate the behavior of our derived acquisition functions.
All relevant details and cross-references are in the captions.

\begin{figure}[!h]
\begin{center}
   \includegraphics{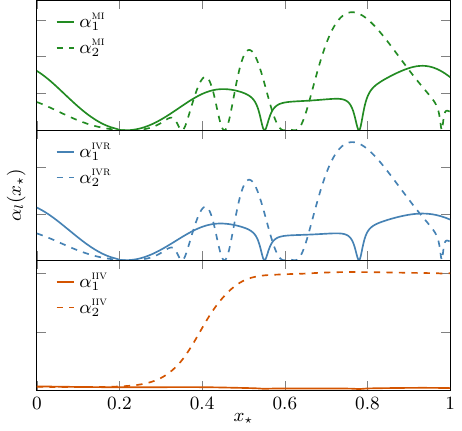}%
\caption{\mi, \ivr, and \ip~acquisitions for the top row of Figure~\ref{fig:demo}. \mi~and \ivr~do not differ a lot, \ie the correlation $\rho$ is rarely large enough for \mi~to leave the linear regime. \mi~puts slightly more emphasis on the primary source where $x_\star$ is close to 1. This indicates that the correlation between $\rz$ and $y_\star$ quite large there. The bottom plot displays the pathology of \ip, where the acquisition for the secondary source essentially follows the inverse cost $c_2$.}
\label{fig:forrester_all_acq}
\end{center}
\end{figure}

\begin{figure}[!h]
\begin{center}
   \includegraphics{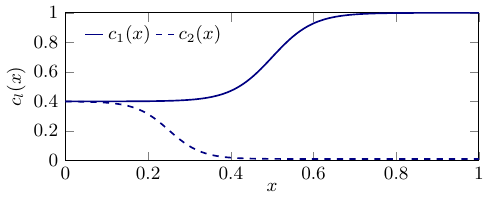}%
\caption{The cost used for the experiment in Figure~\ref{fig:demo}.}
\label{fig:cost-demo}
\end{center}
\end{figure}

\begin{figure}[!h]
\begin{center}
   \includegraphics{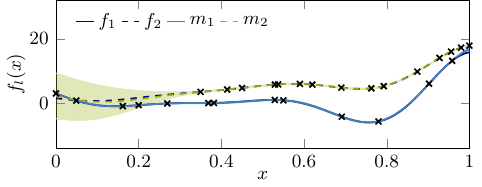}%
\caption{A later state for the experiment shown in Figure~\ref{fig:demo}. Note the absence of $f_2$ evaluations for small $x$ where $c_1$ and $c_2$ are similar.}
\label{fig:finalforrester}
\end{center}
\end{figure}

\begin{figure}[!h]
\begin{center}
   \includegraphics{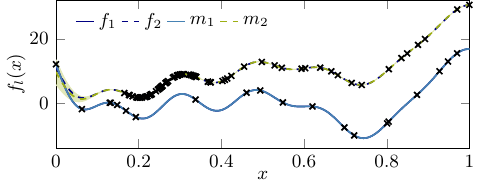}%
\caption{Final state of the \gp~for the second experiment explained in Section~\ref{sec:multiple-sources-non} and shown in Figure~\ref{fig:wiggly} (`wigglified Forrester'). Note the increasing density of evalutions of the secondary source where the cost is minimal, and the lack of $f_2$ queries where $c_1(x) \simeq c_2(x)$. The leftmost evaluation is at the primary source. See Figure~\ref{fig:wiggly} for the cost functions. In this experiment, the \ip~acquisition exclusively evaluates at the location of the minimum of the secondary cost function and is thus stuck.}
\label{fig:finalwiggly}
\end{center}
\end{figure}

\section{DETAILS FOR THE INFECTIONS MODEL}
\label{sec:sirdetails}

\subsection{THE SIR MODEL AND EXTENSIONS}
When the population size is large, the \textsc{sir} (\textsc{s}usceptible, \textsc{i}nfected, \textsc{r}ecovered) model can be described by the following system of ordinary differential equations,
\begin{equation}
\label{eqn:sir}
\begin{aligned}
\frac{\dd\,N_S}{\dd\,t} &= -a \frac{N_S N_I}{N},\\
\frac{\dd\,N_I}{\dd\,t} &= a \frac{N_S N_I}{N} - b N_I,\\
\frac{\dd\,N_R}{\dd\,t} &= b N_I,
\end{aligned}
\end{equation}
in which $a$ is the rate of infection and $b$ the rate of recovery.
It is the most basic of a series of compartmental epidemiological models.
Various extensions exist to accommodate additional effects \eg vital dynamics, immunity, incubation time (cf. \eg \citealp{Hethcote2000}).
Some of these extensions serve as a general model refinement, others are relevant to specific diseases.

Statistical properties, however, are not captured by the description through \textsc{ode}s and call for a stochastic model.
The Gillespie algorithm \citep{Gillespie1976, Gillespie1977} enables discrete and stochastic simulations in which every trajectory is an exact sample of the solution of the `master equation' that defines a probability distribution over solutions to a stochastic equation.
In the \textsc{sir} model, the rate constants are time-independent and thus, the underlying process is Markovian in which the event times are Poisson distributed.
Here, an event denotes the transition of one individual from one compartment to another (\eg $N_I \rightarrow N_R$).
\subsection{EXPERIMENTAL SETUP}

\begin{figure*}[!t]
\begin{center}
    \includegraphics{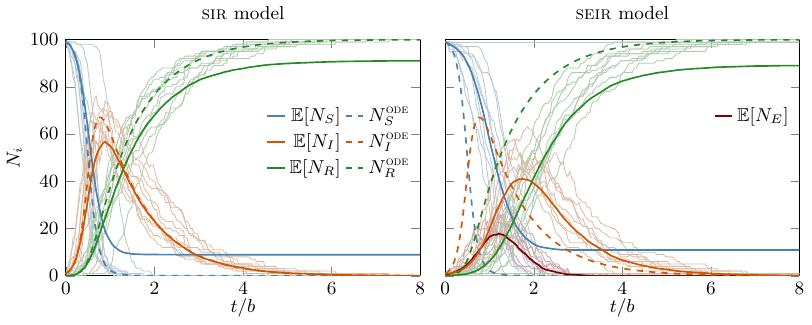}%
\caption{Demonstration of the \textsc{sir} and \textsc{seir} models for $a/b=10$. See text for details.}
\label{fig:sirmodel}
\end{center}
\end{figure*}

For the \amsbq~experiment, we assume that we know the recovery rate $b$, but we are uncertain about the infection rate $a$. 
Therefore, we rescale the \textsc{ode}s and place a shifted gamma prior on $a/b$ that starts at $a/b=1$ and has shape and scale parameters 5 and 4 respectively.
With this prior we encode our belief that the infection rate is significantly larger than the recovery rate so an offset of the epidemic is very likely.
Also, we set the population size to $N=100$ to be well below the thermodynamic limit and set one individual to be infected initially.
We are interested in the expected maximum number $\Exp_a[\max_t N_I(t)]$ of simultaneously infected individuals and the time this maximum occurs $\Exp_a[\argmax_t N_I(t)]$, which might be relevant for vaccination planning.
Querying the primary source $f_1$ for the quantities of interest as a function of $a$ requires numerous realizations of a stochastic four-compartments epidemic model using the Gillespie algorithm \citep{Gillespie1976,Gillespie1977}; in addition to the base model (\textsc{sir}), we include the state `exposed', in which individuals are infected but not yet infectious.
The modified system of \textsc{ode}s that also account for assumed known incubation time $\gamma\inv$ are
\begin{equation}
\label{eqn:seir}
\begin{aligned}
\frac{\dd\,N_S}{\dd\,t} &= -a \frac{N_S N_I}{N},\\
\frac{\dd\,N_E}{\dd\,t} &= a \frac{N_S N_I}{N} - \gamma N_E,\\
\frac{\dd\,N_I}{\dd\,t} &= \gamma N_E - b N_I,\\
\frac{\dd\,N_R}{\dd\,t} &= b N_I,
\end{aligned}
\end{equation}
where we set $\gamma = 10b$.
We absorb the prior on $a/b$ in the black-box function for all methods.

\begin{figure}[!h]
\begin{center}
   \includegraphics{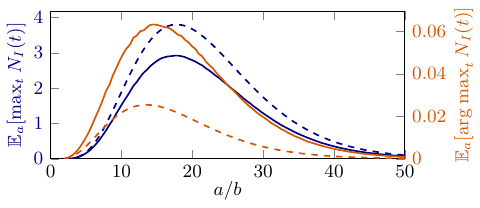}%
\caption{Integrands used for the epidemiological model. Solid lines denote the primary source (\ie stochastic simulations), dashed lines indicate the secondary source (solving the system of \textsc{ode}s). It is apparent from the function that simply integrating the cheap source introduces a significant bias.}
\label{fig:sirintegrands}
\end{center}
\end{figure}

Figure~\ref{fig:sirmodel} shows the \textsc{sir} and \textsc{seir} models (Eq.~\eqref{eqn:sir} and \eqref{eqn:seir}, respectively) with 10 stochastic trajectories (thin lines). 
The solid lines indicate the mean of 100 of these stochastic realizations, and the dashed lines show the solution of the \textsc{ode}s, in both cases for the \textsc{sir} model.
We also use the \textsc{sir} model for solving the \textsc{ode}s even though the stochastic model simulates the \textsc{seir} model.
The purpose of this is to mimic a case where secondary sources are simplified simulations in that minor components are deprecated.
In the stochastic case, there is not always an outbreak of the disease, \ie the initially infected individuum recovers before infecting someone else.
This causes the average $N_R$ to level off significantly below 1.
For the integrals, only outbreaks are taken into account.
The corresponding integrands for the quantities of interest are shown in Figure~\ref{fig:sirintegrands}.

\section{BIVARIATE LINEAR COMBINATION OF GAUSSIANS}
\label{sec:bivar-line-comb}
\begin{figure}[!t]
\begin{center}
  \includegraphics[width=\columnwidth]{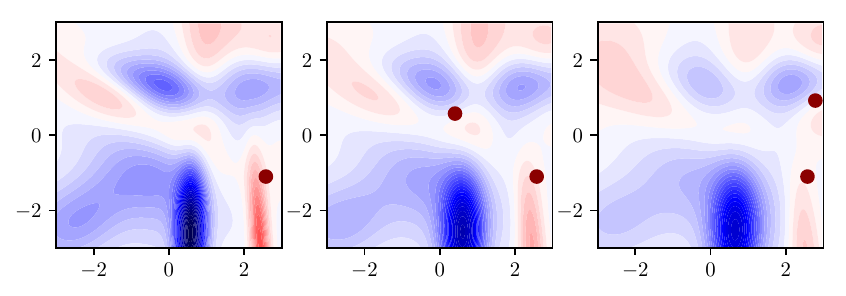}
\caption{Integrands used for the bivariate linear combination of Gaussians. From left to right: primary source $f_1$ and secondary sources $f_2$ and $f_3$. Initial evaluations marked as red dots.}
\label{fig:multigauss_sources}
\end{center}
\end{figure}

We construct an integrand (primary source) $f_1$ in the 2D-domain $[-3, 3]^2$ as a linear combination of $K=20$ normalized Gaussian basis functions 
\begin{equation}
  \label{eq:basis-functions}
\Phi_k^1(\x) = (2 \pi |\sA_k^1|)^{-\frac{1}{2}}e^{-\frac{1}{2}(\x-\m_k^1)\Trans (\sA_k^1)^{-1}(\x-\m_k^1)},
\end{equation} 
i.e., $f_1(\x) = \sum_{k=1}^Kz_k^1\Phi^1_k(\x)$.
 For this, we sample $K=20$ means uniformly $\m_k^1\sim \mathrm{Uniform} [-3, 3]^2$ in the 2D domain. 
We then sample corresponding covariance matices $\sA_k^1$ according to $\u_k^1\sim\N(0, \Id)$, $\boldsymbol{\kappa}_k^1\sim \mathrm{Uniform} [0, 1]^2$, and $\sA_k^1 := \diag(\boldsymbol{\kappa}_k^1) + \u_k^1(\u_k^1)\Trans$. 
The scalar weights $z_k^1$ are sampled from a standard Gaussian $z_k^1\sim \N(0, 1)$ and can be negative. 
Thus $f_1$ is not a probability density function but rather a linear combination of Gaussians with varying location, shape, and weight. 
We then construct secondary sources $f_2$ and $f_3$ consecutively by adding uniform noise to the means, and additive uniform noise to the diagonal of the covariance matrices. 
Thus, with each additional source, each of the $K$ means get randomly but consecutively shifted up and right, and the basis functions $\Phi_k^i(\x)$, $i=2, 3$ randomly become wider and flatter.
Additionally we consecutively add Gaussian random noise to the weights $z_k$ which ensures that the true integrals of the secondary sources differ from the integral of the primary source.
All sources are depicted in Figure~\ref{fig:multigauss_sources}; the primary source $f_1$ on the left, and secondary sources $f_2$ and $f_3$ in the middle and right respectively. 
The cost for evaluating the primary source is $1$ everywhere, the cost of evaluating $f_2$ and $f_3$ are $5\%$ of the primary cost each. 

\begin{figure}[!t]
\begin{center}
   \includegraphics{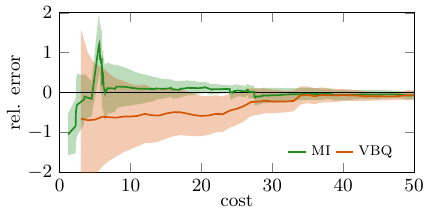}%
\caption{Relative error vs. budget spent for vanilla-\bq~and \amsbq.}
\label{fig:multigauss_result}
\end{center}
\end{figure}

The priors on the kernel lengthscale and coregionalization matrix $\sB$ are set analogously to the other experiments already described in Section~\ref{sec:multiple-sources-non}. 
\amsbq~is initialized with one evaluation of the primary source and two evaluations each of the secondary sources which amounts to a total initial cost of 1.2 (initial evaluations shown as red dots in Figure~\ref{fig:multigauss_sources}). 
Vanilla-\bq~is initialized with three evaluations which are needed to get an initial guess for its hyperparameters (initial cost=3). 
The result is shown in Figure~\ref{fig:multigauss_result} which plots relative error of the integral estimate versus the budget spent as well as two standard deviations of the relative error as returned by the model. 
It is apparent that \amsbq~finds a good solution faster than vanilla-\bq. 

\begin{figure}[H]
\begin{center}
   \includegraphics{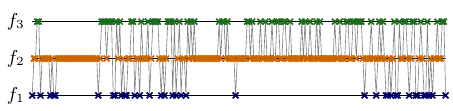}%
\caption{Evaluation sequence of primary and secondary sources in 2D experiment (250 evaluations shown).}
\label{fig:multigauss_chosen_sources}
\end{center}
\end{figure}

Figure~\ref{fig:multigauss_chosen_sources} illustrates the sequence of sources choses by \amsbq.
Secondary source $f_2$ is chosen more often than secondary source $f_3$ at equal evolution cost of 0.05. This is intuitive since $f_2$, by construction, provides more information about $f_1$ than $f_3$, but both secondary sources shrink the budget equally when queried. The percentage of number of evaluations for each source after spending a total budget of 50 is $15\%$, $57\%$, $28\%$ for sources $f_1$, $f_2$, $f_3$ respectively. 

\begin{table*}[!th]
 \begin{center}
    \begin{tabular}{ c c p{10cm}}
    \textbf{Variable} & \textbf{Shape} & \textbf{Description}\\
    \hline
    $L$     && number of sources, indexed by $l$ where $l=1$ is the primary source \\
    $D$     && dimension of the input space, indexed by $d$\\
    $N$     && number of source-input-evaluation triplets, indexed by $n$ \\
    $N_\star$     && number of potential new source-input-evaluation triplets\\
    $\x$    & $D\times 1$ & input location\\
    $\fvec(\x)$ & $L\times 1$ & $[f_1(\x)\ \dots\ f_L(\x)]\Trans$, where $f_l(\x)$ is the $l^\mathrm{th}$ source\\
    $\intangle{f_l}$  && $\int_\Omega f_l(\x)\, \dd \pi(\x)$ integral of the $l^\text{th}$ source\\
    $\pi(\x)$ && integration measure on $\Omega$\\
    $\Omega$  & & domain that is integrated over, $\Omega \subseteq \Re^D$\\
    $\rz$ & & random variable for integral of interest $\rz\sim \N (\Exp[\rz\g\Dcal], \V[\rz\g\Dcal])$\\
    $(l_n, \x_n)$  & $(1, D\times 1)$ & $n^\mathrm{th}$ source-location pair where $\fvec$ is evaluated\\
    $(\ldat, \sX)$   & $(N\times1, N\times D)$ & $N$ source-location-pairs $([l_1\ \dots\ l_N]\Trans, [\x_1\ \dots\ \x_N]\Trans)$\\
    $\f_{\ldat}$   & $N\times 1$ & $ [f_{l_1}(\x_1)\ \dots\ f_{l_N}(\x_N)]\Trans$ noise-free function evaluations\\
    $\y_{\ldat}$   & $N\times 1$ & $ [f_{l_1}(\x_1) + \eps_{l_1}\ \dots\ f_{l_N}(\x_N) + \eps_{l_N}]\Trans$ noisy function evaluations\\
    $\vec{y}$ & $L\times 1$ & $\vec{y} = \fvec(\x) + \vec{\eps}$ simultaneous evaluation of all sources\\
    $\vec{\eps}$ &$L\times 1$& $\vec{\eps} = [\eps_1\ \dots\ \eps_L]$ noise vector, $\eps_l\sim\N (0, \sigma_l^2)$\\
    $\sSigma_{\ldat}$ & $N\times N$ & $= \diag (\sigma_{l_1}^2, \dots, \sigma_{l_N}^2)$ diagonal noise matrix with noise per level $\sigma_{l_n}^2$\\
    $\Dcal$ & & $N$ collected data triplets $\{(l_n, \x_n, f_{l_n} (\x_n))\}_{n=1}^N$\\
    $\sK$  & $L\times L$ & $\sK = \cov[\fvec, \fvec]$ matrix-valued covariance matrix\\
    $k_{ll'}(\x, \x')$ && covariance function $\cov [f_l(\x), f_{l'}(\x')]$\\
    $\kvec_{l\ldat} (\x, \sX)$ & $1\times L$ & vector-valued covariance $\cov [f_l(\x), \f_{\ldat}(\sX)]$\\
    $\sK_{\ldat \ldat} (\sX,\sX)$ & $N\times N$ & $\cov [\f_{\ldat}(\sX), \f_{\ldat}(\sX)]$\\
    $\sG_{\ldat} (\sX)$ & $N\times N$ & Gram matrix $\sK_{\ldat \ldat} (\sX, \sX) + \sSigma_{\ldat}$\\
    $\m (\x)$ & $L\times 1$ & \gp~prior mean for multi-output \gp\\
    $\m_{\ldat} (\sX)$ & $N\times 1$ & prior mean evaluated at source-location pairs $(\ldat, \sX)$ \\
    $m_{l \g \Dcal} $ && posterior mean at source $l$\\
    $k_{ll' \g \Dcal} $ && posterior covariance of sources $l, l'$\\
    $\intangle{m_l}$ && $\int_\Omega m_l (\x)\,\dd \pi(\x)$ integrated prior mean\\
    $\intangle{\kvec_{\ldat l} (\sX, \cdot)}$ & $1\times L$ & kernel mean of $l^{\text{th}}$ source at source-location pairs $(\ldat, \sX)$ \\
    $\iintangle{k_{ll'}}$ && $\iint_\Omega k_{ll'} (\x,\x')\, \dd \pi(\x) \dd\pi(\x')$ initial error\\
    $\sB$ & $L\times L$ & coregionalization matrix for the kernel used in the \textsc{icm}\\
    $\kappa(\x, \x')$ & & kernel encoding purely spatial correlation in the \textsc{icm}\\
    $(\ldat_\star, \sX_\star, \y_{\ldat_\star})$ & $(N_\star\times 1, N_\star\times D, N_\star\times 1)$ & potential new source-location-evaluation triplets\\
    $c_{\ldat_\star} (\sX_\star)$ && cost of evaluating at $(\ldat_\star, \sX_\star)$; $c_{\ldat_\star} (\sX_\star) = \sum_{i=1}^{N_\star} c_{l_i} (\x_i)$\\
    $\sV_{\ldat_\star | \Dcal} (\sX_\star)$ & $N_\star\times N_\star$ & $= \sK_{\ldat_\star\ldat_\star | \Dcal}(\sX_\star, \sX_\star) + \sSigma_{\ldat_\star}$; in the myopic case denoted as $v_{l_\star | \Dcal} (\x_\star)$ \\
    $\rho^2_{1 \ldat_\star | \Dcal} (\sX_\star)$ && scalar correlation function for $(\ldat_\star, \sX_\star)$, defined in Eq.~\eqref{eqn:rho2}\\
    $\alpha_{\ldat_\star} (\sX_\star)$ && non-myopic acquisition function, $\alpha_{l_\star} (\x_\star)$ in the myopic case\\
    \end{tabular}
\end{center}
\caption{Summary of the notation used. Generally, vector-valued quantities are denoted by lower case bold letters and matrices are upper case bold letters. Normal font denotes scalars.}
\label{tbl:notation}
\end{table*}

\end{document}